\documentclass[journal,onecolumn]{IEEEtran}
\IEEEoverridecommandlockouts
\usepackage{cite}
\usepackage{amsmath,amssymb,amsfonts}
\usepackage{algpseudocode}
\usepackage{graphicx}
\usepackage{textcomp}
\usepackage{xcolor}
\usepackage[pagebackref,breaklinks,colorlinks]{hyperref}
\usepackage{caption}
\usepackage{subcaption}
\usepackage{multirow}
\usepackage[bottom]{footmisc}
\usepackage{algorithm}
\usepackage{listings}

\definecolor{codegreen}{rgb}{0,0.6,0}
\definecolor{codegray}{rgb}{0.5,0.5,0.5}
\definecolor{codepurple}{rgb}{0.58,0,0.82}
\definecolor{backcolour}{rgb}{0.95,0.95,0.92}

\lstdefinestyle{mystyle}{
    backgroundcolor=\color{backcolour},   
    commentstyle=\color{codegreen},
    keywordstyle=\color{magenta},
    numberstyle=\tiny\color{codegray},
    stringstyle=\color{codepurple},
    basicstyle=\ttfamily\footnotesize,
    breakatwhitespace=false,         
    breaklines=true,                 
    captionpos=b,
    keepspaces=true,                 
    numbers=left,                    
    numbersep=5pt,                  
    showspaces=false,                
    showstringspaces=false,
    showtabs=false,                  
    tabsize=2
}

\newcommand{\R}{\mathbb{R}}

\usepackage[capitalize]{cleveref}
\crefname{section}{Sec.}{Secs.}
\Crefname{section}{Section}{Sections}
\Crefname{table}{Table}{Tables}
\crefname{table}{Tab.}{Tabs.}

\def\BibTeX{{\rm B\kern-.05em{\sc i\kern-.025em b}\kern-.08em
    T\kern-.1667em\lower.7ex\hbox{E}\kern-.125emX}}
\begin{document}

\title{\LARGE LAP: An Attention-Based Module for Faithful Interpretation and Knowledge Injection in Convolutional Neural Networks
}

\author{\IEEEauthorblockN{Rassa Ghavami Modegh\textsuperscript{1},
Ahmad Salimi\textsuperscript{1},
Alireza Dizaji\textsuperscript{1},
Hamid R. Rabiee\textsuperscript{*,1}
\thanks{\textsuperscript{*}Corresponding author: \href{mailto:rabiee@sharif.edu}{rabiee@sharif.edu}}}\\
\IEEEauthorblockA{
\small
\textsuperscript{1}\textit{Department of Computer Engineering, Sharif University of Technology, Tehran, Iran}
}
}

\maketitle

\begin{abstract}
Despite the state-of-the-art performance of deep convolutional neural networks, they are susceptible to bias and malfunction in unseen situations. Moreover, the complex computation behind their reasoning is not human-understandable to develop trust. External explainer methods have tried to interpret network decisions in a human-understandable way, but they are accused of fallacies due to their assumptions and simplifications. On the other side, the inherent self-interpretability of models, while being more robust to the mentioned fallacies, cannot be applied to the already trained models. In this work, we propose a new attention-based pooling layer, called Local Attention Pooling (LAP), that accomplishes self-interpretability and the possibility for knowledge injection without performance loss. The module is easily pluggable into any convolutional neural network, even the already trained ones. We have defined a weakly supervised training scheme to learn the distinguishing features in decision-making without depending on experts' annotations. We verified our claims by evaluating several LAP-extended models on two datasets, including ImageNet. The proposed framework offers more valid human-understandable and faithful-to-the-model interpretations than the commonly used white-box explainer methods.
\end{abstract}


\section{Introduction}
\label{sec:intro}

Nowadays, Artificial Intelligence (AI) has entered into real-life applications like clinical computer-aided decision systems, medical diagnosis, and autonomous car driving. These critical applications are concerned about whether AI models are trustable and whether their decisions are valid \cite{tjoa2020survey}. Deep Neural Networks (DNNs), one of the most successful AI models, make decisions using complex computations humans do not understand. They are trained end-to-end and are susceptible to learning detours and biases of the dataset rather than the actual concepts and reasons. Since AI has become responsible for making decisions in areas interfering with human rights and ethics, governments have started to make laws about its usage. For example, the European Union has adopted new regulations that enable users to demand an explanation of an algorithmic decision that has affected them \cite{goodman2017european}. This has strengthened the urge for DNNs to explain themselves. Explaining DNNs has other virtues besides verifying decisions, bias detection, developing trust, and compliance to legislation \cite{caruana2015intelligible}; it can help diagnose the model. Also, knowledge can be discovered from the models with superior-than-human performance to enrich human knowledge \cite{du2019techniques}.
\par
In recent years, there have been many attempts to explain and interpret DNNs' decisions. Global interpretation methods focus on explaining the overall decision-making process of models to develop trust and prove unbiasedness. On the contrary, local interpretation methods tend to explain the decision-making process for a single sample, which helps validate single decisions. From another point of view, interpretation methods can be divided into two general areas of intrinsic and post-hoc methods. Intrinsic interpretability is achieved by enforcing interpretability into the model's architecture \cite{sabour2017dynamic,wu2017towards,zhang2018interpretable} and the training strategy \cite{kingma2013auto,chen2016infogan,freitas2014comprehensible,zhang2018interpretable}. In this approach, the model itself can provide explanations for its decisions. These methods do not apply to already trained models \cite{gilpin2018explaining,kim2018interpretability}. They generally pose limitations over the model's architecture, and some may sacrifice the performance to achieve interpretability \cite{gilpin2018explaining}. Post-hoc methods try to provide explanations for already trained models. They adopt assumptions and simplifications in their computations that may sacrifice fidelity to achieve more human-understandable explanations \cite{gilpin2018explaining}. These assumptions also may lead to false interpretations as they may not be valid in the model's decision-making process \cite{sixt2020explanations}.
\par
Recently, a new group of global interpretation methods has emerged. These methods aim to find the global, human-understandable concepts that are effective in the overall decision-making process. In the literature on these models, concepts are defined as semantically meaningful groups of input features. The concepts have been preferred to raw single input features as they are more understandable \cite{schwalbe2022concept}. Several studies have utilized expert-defined task-related concepts annotated in the whole dataset. Some studies focusing on post-hoc global interpretation, like TCAV \cite{kim2018interpretability} and NBR \cite{graziani2018regression}, have measured the sensitivity of a task with respect to the expert-defined concepts as a measure of the validity of the model's performance. Some other studies have focused on intrinsically interpretable groups of models named Concept Bottleneck Models (CBMs) \cite{koh2020concept}. They break the task into two parts: finding the existence of the defined concepts and making the final decision based on the output of the first part. These studies aim to ensure the users that the model makes the right decision based on the right reasons. They also allow the intervention of humans by correcting the found concepts directly at the test time \cite{koh2020concept}. While the models guarantee validity, the concept finding submodel remains nontransparent. CBMs require expert knowledge about the task and are also susceptible to performance drop due to incompleteness of the defined concept set or insufficiency of the final decider in modeling inter-concept interactions. Besides, they do not allow the models to learn beyond the existing knowledge of the problem. But their biggest challenge is the scarcity and expensiveness of richly densely annotated samples \cite{schwalbe2022concept}. To deal with the issues of expert-defined concepts, some studies have focused on mining the concepts after training using unsupervised methods like super-pixel semantic segmentation, clustering \cite{ghorbani2019towards}, matrix factorization, and principal component analysis \cite{zhang2021invertible}. Some studies measure the completeness of the defined or mined concept set as the performance of a distilled model working purely based on the concepts \cite{schwalbe2022concept, zhang2021invertible}. While the mentioned literature has focused on global interpretation for validating the model, some studies have utilized similar terms seeking different objectives. HINT \cite{wang2022hint} has used existing post-hoc explainers alongside the hierarchical relations between the class labels in a classification task to train a  distilled model for whole object capturing in a weak-supervised object segmentation task.
\par
In this work, we have designed a novel module called Local Attention Pool (LAP), a concept system, and a training scheme to enhance the interpretability of the models. LAP is easily pluggable into any CNN architecture, providing intrinsic interpretability. It also can be used as an external module for distilled model training, providing post-hoc interpretability. LAP is responsible for highlighting the concepts required for the task. Despite the concept-based interpretation literature, we have employed concepts to provide visual local or instance-based interpretations. Our training scheme encourages the model to detect the concepts explicitly while training. Although the concept system and the training scheme can be used in a supervised fashion using experts' defined concepts and annotations, they do not necessarily depend on that. We have provided guidelines for classification as an example task to allow a non-expert to design necessary concepts and employ the training scheme without concept annotations to encourage the model toward detecting the concepts on its own. In this design, a concept is a distinguishing feature that separates two mutually exclusive and collectively exhaustive subsets of categories. For instance, "having wings" is a concept that separates images of "birds" and "airplanes" from those of "cats" and "cars" in a 4-class classification. To achieve pluggability, we designed LAP as a pooling layer that can replace pooling layers and strides in CNN architectures. The reason is that only pooling layers and strides act as stream selection, and replacing them with another stream selection must not hurt the performance of an already-trained network. Although LAPs can be used as an external module, to test the performance preserving, we compared the interpretability of plugged-in LAP extended models with other local interpretation methods on two settings, training from scratch and tuning an already trained architecture.
We showed that LAPs can detect distinguishing features better than the other methods without causing a performance drop. Our main contributions can be summarized as follows:
\begin{itemize}
    \item Introducing a new module that is easily pluggable to any CNN, including the already trained networks, that accommodates the model with self-interpretability and the possibility to inject knowledge without restricting the architecture, performance loss, and adding parameters to the main-stream of information flow (so not sensible to overfitting more than the base model). 
    \item Proposing a concept-wise attention mechanism that assigns attention scores to any predefined domain concepts to distinguish the importance of each pixel according to each concept. 
    \item Proposing a weakly supervised method for encouraging the model toward explicit detection of the concepts without depending on detailed concept labels, resulting in enhanced interpretability.
\end{itemize}

\section{Background}
\label{sec:related}

\subsection{Local Interpretability methods}

In recent years, many studies have been published about the interpretability and explainability of DNNs. Some of the proposed methods are model-agnostic and treat models as black-boxes. One group of model-agnostic methods mimics the operations of the black-box by training a white-box model and interpreting it instead \cite{vandewiele2016genesim}, which is susceptible to errors as the mimicked model does not perform exactly as the primary model. Another group of model-agnostic methods like LIME \cite{ribeiro2016should} assesses feature sensitivity by perturbing the feature space around each input, which demands an optimization per sample and is computationally inefficient for being applied on many samples with large feature spaces like images.
\par
Model-specific methods work on specific white-box models, meaning they use the architecture and parameters of the model to provide explanations. CAM \cite{zhou2016learning} only applies to the networks having one final fully connected layer after the last convolutional layer. It uses the weights of features in the fully connected layer to find the importance of the features for each class and then calculates pixel scores based on their channel-wise activations and channel importance. Gradient-based methods use gradients of class scores w.r.t the input to find the most sensitive features as if the network is estimated with a Taylor series of order one. The gradients are the weights of the features indicating their importance. Vanilla gradient \cite{simonyan2013deep} uses pure gradients to find important features. It produces a noisy importance map. Guided Backpropagation \cite{springenberg2014striving} filters negative gradient flows to bold out the effective pixels. The filtering may lead to false positives. Grad-CAM \cite{selvaraju2017grad} follows the same steps as CAM, but it uses gradients instead to find the importance of channels, making it applicable to a broader group of networks. Guided Grad-CAM \cite{selvaraju2017grad} is a multiplication of Guided Backpropagation and Grad-CAM to produce fine-grained importance maps.
\par
The gradient-based methods suffer from gradient saturation problems, leading to near-zero importance scores. Score-based methods like Layer-wise Relevance Propagation (LRP) \cite{bach2015pixel, binder2016layer} and Deep Lift \cite{shrikumar2017learning} propagate scores instead of gradients to calculate the importance of neurons. LRP has defined layer-specific rules to divide the relevance score of the neurons in each layer to their input neurons. The rules assign relevances based on the contribution of the input neurons to the neurons in the next layer. Deep Lift uses a similar procedure to LRP but divides the scores based on the difference in the output achieved compared to a baseline input. However, defining a suitable and meaningful baseline for all applications is challenging. Recently, some works have combined the CAM method with score-based interpretations to improve both. Score-CAM \cite{wang2020score} adopts a Deep Lift-style scoring scheme to find the channel-wise increase of confidence and then uses the confidence scores in the CAM method. Similar to Score-CAM, Relevance-CAM \cite{lee2021relevance} uses the LRP method to find the importance of channels in any layer and then applies the CAM method to find the corresponding regions in the input. Relevance-CAM has proved its superior performance to other CAM-based methods. 
\par
Another group of methods has used attention maps generated by attention mechanisms to explain the model's behavior. This kind of explanation is widely used on Vision Transformers \cite{vit} in various domains, e.g., pose estimation \cite{yang2021transpose} and medical image diagnosis \cite{mondal2021xvitcos}. A field of research is focused on enhancing transformer explanations by introducing attention flow \cite{abnar2020rollout} and relevancy propagation \cite{Chefer_2021_CVPR, Chefer_2021_ICCV}  through the layers. There has also been an attempt to make CNN models interpretable by adopting 
self-attention layers \cite{gu2020net, Zhang2021alzheimer}. These attention-based explanation methods only apply to specific architectures they proposed or the Transformers. Therefore, they differ from our work, which applies to any CNN architecture.

\subsection{Importance based pooling}

Pooling layers and strided convolutions are widely used in CNNs to increase the receptive field and decrease memory consumption. Gao et al. have proposed a unified framework for formulating different pooling strategies, called Local Aggregation and Normalization (LAN), and a pooling method called Local Importance Pooling (LIP) \cite{gao2019lip}. This framework aggregates features within local sliding windows by weighted averaging. The weights are assigned based on the importance of the features. According to this framework, average pooling assumes the same importance score for all the pixels and is susceptible to feature fading. Max pooling assigns one to the highest feature and zeroes to all the others, leading to sparse gradient paths and slow training. Strided convolutions give importance based on the pixel's location in the window and are more sensitive to shift variances. LIP has used the attention mechanism to assign importance weights to the features. LIP is applied feature-wise, which makes it different from our proposed architecture. All of the mentioned pooling layers except strided convolutions can lead to loss of relative spatial relations, as they select different features from different spatial locations \cite{patrick2019capsule}.

\section{Method}
\label{sec:method}

\subsection{Local Attention Pool}
\begin{figure*}[htb]
\centering
\subfloat[]{\includegraphics[width=0.495\textwidth]{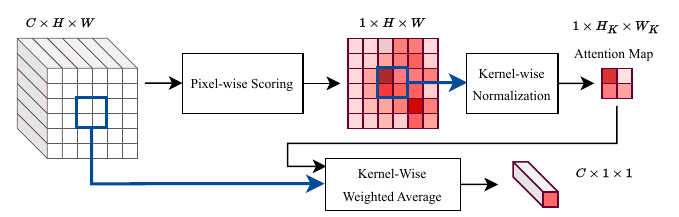}%
\label{fig:attention:mech}}
\hfil
\subfloat[]{\includegraphics[width=0.495\textwidth]{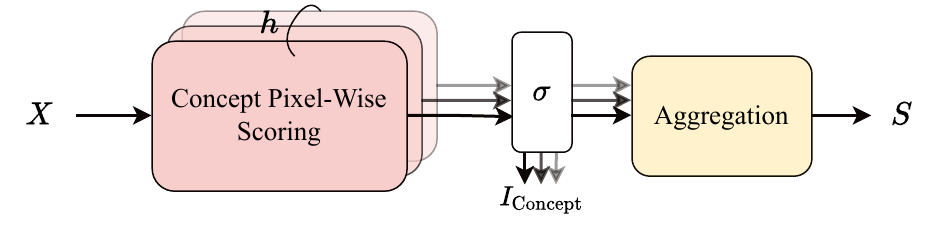}%
\label{fig:mhattention}}
\caption{(a) The local attention pooling is applied on a kernel of size $W_K \times H_K$ on the input feature map. (b) The multi-concept pixel-wise scoring module aggregates pixel-wise importance maps calculated for all concepts, $I_{\text{concept}}$, into a single score map, $S$.}
\label{fig:attention}
\end{figure*}

The attention mechanism generally conveys information about the most important parts of the input data. We adopt the attention mechanism in the reduction process of the pooling layers. The process is depicted in \cref{fig:attention}. In contrast to LIP, attention is not applied in a channel-wise manner. Instead, the whole feature map is passed to a scoring module to calculate pixel-wise importance scores. Then, the scores related to the pixels under the kernel are normalized for each kernel position. The final feature vector is obtained by weighted averaging the feature vectors of the pixels. This process mimics a zooming action. Instead of mixing the features of the pixels applied to a pooling kernel, the network dynamically detects the most important pixels and passes their features instead. This way, the small yet important details are not faded or lost but propagated through the network's depth. It also prevents fallacies produced by feature mixing. Different zooming locations under each kernel position may also help make the model more robust toward shifts and scales.

We have modified the LAN framework to express the pooling procedure of LAP in \cref{eq:lanform}. In this equation, the output $O_{i', j'}$ corresponding to a sliding kernel of size $W_K \times H_K$ with the top-left corner of $(i, j)$ is calculated based on the input feature map $X$ related to the pixels under the kernel, $X_{i:i+H_K,j:j+W_K}$, and the weighting function $F$. In this equation, $\odot^*$ stands for element-wise multiplication with casting, as a three-dimensional tensor $X_{i:i+H_K,j:j+W_K} \in \mathbb{R}^{C \times H_K \times W_K}$ is being multiplied by a two-dimensional tensor $F(X_{i:i+H_K,j:j+W_K}) \in \mathbb{R}^{H_K \times W_K}$. The problem with LAN was its calculation of importance weights based on single values, while LAP uses the whole feature map under a kernel to calculate the weights.

\begin{equation}
\left\{
  \begin{aligned}
      & 
       O_{i',j'}=\frac{\Sigma (F(X_{i:i+H_K,j:j+W_K}) \odot^* X_{i:i+H_K,j:j+W_K})}{\Sigma F(X_{i:i+H_K,j:j+W_K})}
      \\
      & 
      F: \mathbb{R}^{C \times H_K \times W_K} \to \mathbb{R}^{H_K \times W_K}
  \end{aligned}
\right.
  \label{eq:lanform}
\end{equation}

In this work, $F$ combines two steps: pixel-wise scoring, $S$, and local kernel-wise normalization, $N$. These steps are presented in \cref{fig:attention}. The general form of $F$ can be illustrated as \cref{eq:lapformal} in which $S$ and $N$ can be any arbitrary function.

\begin{equation}
\left\{
\begin{aligned}
&F(X_{i:i+H_K,j:j+W_K}) = N(V_{i:i+H_K,j:j+W_K})
\\
&V_{a,b} = S(X_{a,b}) \\
&N: \mathbb{R}^{H_K \times W_K} \to \mathbb{R}^{H_K \times W_K}; S: \mathbb{R}^{C} \to \mathbb{R}
\end{aligned}
\right.
\label{eq:lapformal}
\end{equation}
For deep models to effectively perform their tasks, they must learn one or more problem-specific concepts. To embed this process into the model's architecture, we designed a pixel-wise scoring module $S$ as depicted in \cref{fig:mhattention}. In this design, we have considered $h$ concept scoring heads, each responsible for assigning an importance score to the pixels for their corresponding concept. Interpretability is one of the well-known benefits of the attention mechanism. Attention scores identify the relative importance of each pixel of the input. Any arbitrary function can be used for scoring the pixels. The naive scores require binarization to denote the "important pixels" in an absolute sense rather than expressing relative importance. To address this issue, we convert the concept-wise importance scores into concept-wise importance probabilities, denoted as $I_{\text{Concept}}$, by applying a sigmoid function. Using these probabilities allows us to apply losses for classifying importance, which is further explained in \cref{subsec:ki}. The final pixel-wise importance score is calculated by aggregating the concept-wise importance probabilities. The process can be formulated as \cref{eq:mhattention}, where $\sigma$ is the sigmoid function, $S_C$ is the concept pixel-wise scoring function, and $A$ is the aggregation function. While $S_C$ is a trainable module, $A$ can either be a trainable or fixed aggregation function, such as maximum.
\begin{equation}
\begin{aligned}
&S(x) = A(\sigma(S_C(x)))\,;\, S_C: \mathbb{R}^C \to \mathbb{R}^h, A: \mathbb{R}^h \to \mathbb{R}
\end{aligned}
\label{eq:mhattention}
\end{equation}

One can adopt any arbitrary method to locally normalize the importance scores based on the pixels' scores under the kernel at each position. In this study, we have adopted weighted averaging for kernel normalization as presented in \cref{eq:gkernel}. The weights are calculated based on adjusting the final importance scores using a Gaussian kernel around the highest local importance probability. Therefore, the sensitivity toward the most highlighted local pixel becomes adjustable by the trainable parameter $\alpha$. Theoretically, when $\alpha = 0$, $W(V) = 1 + \epsilon$, and as $\alpha \to \infty$, the coefficient of the pixels other than $\max\{V\}$ would approach $\epsilon$. LAP would convey the features of the most highlighted local pixel directly. We have added the small value $\epsilon$ to prevent the weights from becoming zero. This value helps preserve the gradient flow to all pixels and prevents zero division in the weighted averaging process.

\begin{equation}
\left\{
\begin{aligned}
& N (V_{i:i+H_K,j:j+W_K}) = \frac{W^{i,j}_{i:i+H_K,j:j+W_K} \odot V_{i:i+H_K,j:j+W_K}}{\Sigma (W^{i,j}_{i:i+H_K,j:j+W_K})}\\ 
& W ^{i,j}_{a,b} = e^{-\alpha ^2 (\max({\{V_{i:i+H_K,j:j+W_K}\}}) - V_{a, b})^2} + \epsilon \\
& \forall{a, b}; \quad i \leq a < i + H_K, j \leq b < j + W_K.
\end{aligned}
\right.
\label{eq:gkernel}
\end{equation}

CNNs are generally a stack of layers, most of which work by sliding a kernel all over the input and calculating a function. Although the shallower layers have lower receptive fields, they extract common details like edges and corners. As the depth increases, the receptive field increases, and the layers become responsible for extracting higher-level concepts \cite{olah2017feature}. The final decision is made based on the information flow through the layers, and somewhere in the middle, it should have understood the distinguishing concepts. LAP helps specify those concepts besides keeping them bold in the pooling processes. As the importance is identified based on the internal forward flow of the network, LAP modules make the model self-interpretable. LAPs do not produce false positives or negatives due to ignored flows to reach a human-understandable interpretation in contrast to external explainer methods. They also find the important parts in the same direction the model decides in the forward process.

\subsection{Knowledge injection}
\label{subsec:ki}

Human experts make their decisions based on special features of the input. For example, jaggedness is one of the factors considered in classifying a tumor as benign or malignant. The neural networks trained freely may or may not have considered all the reasons in their decision-making. They may have become biased toward a dominant feature in the training dataset and lose generalization. LAP provides an easy way to inject experts' knowledge into the network due to the probabilistic behavior of its scoring module. Experts can highlight the important input parts for their decision-making for each concept. The highlighted map can be resized to each LAP's input and used as ground truth to train each concept scoring head. Knowledge injection gives a better guarantee that the network decides based on the factors of the domain.
\par
In most situations, we do not have detailed experts' supervision, but we have general knowledge about the problem. So, the LAP modules can be trained weakly supervised to behave as desired. Therefore, it will also benefit from gradient injection and fast training. We have used a linear combination of the semi-supervised losses and the main task's loss to train the models.

\subsubsection{Concept-discrimination loss}

For adopting the mentioned loss, we used the design presented in \cref{fig:mhattention} for the scoring module, with $h$ concept heads. We assume each sample $s$ has a set of concepts $C_s$. Each concept head highlights the important pixels for the concept, called concept-related clues. To train the concept heads, we considered the following loss terms:

\noindent\textbf{Min Active Ratio (MinAR).} Each concept head should assign high importance probability to at least a specified portion of the pixels in the samples containing the concept. This makes the model highlight the concept-related clues for the sample.

\noindent\textbf{Max Active Ratio (MaxAR).} Each concept head should assign high importance probability to, at most, a specified portion of the pixels in the samples containing the concept, as the main clue is not understandable otherwise. This phenomenon may happen in layers with large receptive fields. As the concept-related clue appears in the receptive fields of all pixels, the network may consider them concept-related clues.

\noindent\textbf{Inactive Ratio (IAR).} Each concept head should assign low importance probability to all the pixels of the samples not containing the concept, as the concept-related clue should not exist in them. This loss can be applied to all the pixels. Nevertheless, since most of the pixels in the attention map are already inactive, this loss term will likely fade over time. Therefore, applying this term on top-scored pixels is better according to IAR.

Consider a LAP module $l$ with an input size $H \times W$. The loss function for this LAP is shown in \cref{eq:mcloss}. In this equation, $N_c$ and $N_{\hat{c}}$ are the number of samples containing and not containing concept $c$, respectively. $k_1 = \lceil \text{MinAR} \times HW \rceil $ and $k_2 = \lceil \left(1 - \text{MaxAR}\right) HW \rceil$ are the numbers of pixels encouraged to have respectively high and low probabilities in the samples containing concept $c$, and $k_3 = \lceil \text{IAR} \times HW \rceil$ is the number of top pixels encouraged to be inactive in other samples. $\text{top}_{s,k_1}^{l,c}$ and $\text{bot}_{s,k_2}^{l,c}$ are the sets of pixels of the $k_1$ high-ranked and $k_2$ low-ranked pixels of sample $s$ based on the importance probability of concept head $c$. $p_{s, i, j}^{l, c}$ is the probability of concept head $c$ for the pixel $(i, j)$ of sample $s$. MinAR, MaxAR, and IAR are hyper-parameters.  The first term is multiplied by $2$ to balance the effect of positive and negative losses to the concept head $c$.

{
\begin{equation}
      -\sum_{c=1}^h \Bigg[ \sum_{s;c \in C_s} \bigg[
       \frac{2 \sum_{(i,j) \in \text{top}_{s,k_1}^{l,c}} \ln(p_{s,i,j}^{l,c})}{k_1 \times N_c} 
      + \frac{
      \sum_{(i,j) \in \text{bot}_{s,k_2}^{l,c}} \ln(1 - p_{s,i,j}^{l,c})}{k_2 \times N_c} 
      \bigg]
      +\sum_{s;c \notin C_s} \bigg[ \frac{
      \sum_{(i,j) \in \text{top}_{s,k_3}^{l,c}} \ln(1 - p_{s,i,j}^{l,c})}{k_3 \times N_{\hat{c}}}
      \bigg] \Bigg]
  \label{eq:mcloss}
\end{equation}
}

In our experiments, we observed that choosing $\text{top}_{s,k}^{l,c}$ and $\text{bot}_{s,k}^{l,c}$ sets based on the probabilities assigned by concept head, is sensitive to the initial model parameters. If high weight is considered for concept-discrimination loss, it will likely get stuck considering a wrong zone in layers with high receptive fields. To prevent this, we used another module with the same architecture as the scoring module to choose $\text{top}_{s,k}^{l,c}$ and $\text{bot}_{s,k}^{l,c}$ sets. The module, called the discriminative scoring module, is trained using the first and third terms of the loss function presented in \cref{eq:mcloss}, without multiplying the first term by 2, and with $k_1 = k_3 = HW$. We detached the input of this module to prevent misleading loss injection to the model, as many input parts may be common between the concepts.

\subsubsection{Knowledge sharing by concordance loss}

Intuitively, if one LAP layer has found a part of the input as the distinguishing clue for one concept, the proceeding LAP layers should also distinguish the same clue. This fact does not always hold for the previous layers, as they might not understand enough to perceive the clue, mainly due to the low receptive field. A human expert can decide whether the receptive field is enough for the LAP layers. In that case, we have used the Jensen-Shannon divergence loss to encourage the consecutive LAP layers to produce similar maps. The loss is presented in \cref{eq:js}. $\mathcal{JS}(l, s, c)$ stands for the loss for the concept head $c$ of sample $s$ between the LAPs $l$ and $l + 1$, and $M$ is the total number of pixels.

{
\begin{equation}
      \mathcal{JS}(l, s, c) =
      \frac{1}{2M}\sum_{(i,j)}
      \bigg[
      (p_{s,i,j}^{l,c} - p_{s,i,j}^{l+1,c})\ln{\frac{p_{s,i,j}^{l,c}}{p_{s,i,j}^{l+1,c}}} +
      (p_{s,i,j}^{l+1,c} - p_{s,i,j}^{l,c})\ln{\frac{1 - p_{s,i,j}^{l,c}}{1 - p_{s,i,j}^{l+1,c}}}
      \bigg]
  \label{eq:js}
\end{equation}
}

Using the Jensen-Shannon loss causes the LAP layers to help each other in finding more clues, and the found clues are also more robust. If the receptive field does not suffice in some layer, e.g., $l$, only the pixels with high importance probabilities of $l$ and low importance probabilities of $l + 1$ can be used in the Jensen-Shannon divergence loss. We have applied this loss between each pair of consecutive LAP layers. 

\subsection{LAP-Extended Models}

LAP is easily pluggable into any convolutional architecture. Pooling and adaptive pooling layers can be replaced directly with LAPs. Strided convolutions can also be replaced by a convolution with the stride of one, proceeding with a LAP with the same kernel size and stride as the convolution's stride. The unique advantage of LAP is that it can be plugged into an already-trained model and tuned while other model layers are frozen.

\section{Results}

In the experiments, we aimed to show the general applicability of LAP to different domains and architectures without performance loss and compare its interpretations with state-of-the-art methods for interpreting a model locally with no or slight modification. Following the objective, we have not compared the interpretation method with methods not applicable to all CNNs in general (e.g., the transformers), the ones requiring training from scratch (e.g., black-box explainers and other self-interpretation methods), the ones that have not provided an algorithm for a unified interpretation (e. g. LAN), and the methods that depend to other explainers we have compared our results with (e.g., HINT \cite{wang2022hint}). Also, the human-centered interpretation methods are set aside as they need a human for the task. We compared our interpretations with five local white-box explainer methods, Guided Back-propagation (GBP), Grad-CAM (GC), Guided Grad-CAM (GGC), Deep Lift (DL), and Relevance-CAM (RC), using implementations of captum \cite{kokhlikyan2020captum} in  PyTorch \cite{NEURIPS2019_9015}. Notably, the selected methods are among the most widely used white-box explainers, and they cover three different strategies, score back-propagation, gradient-based, and CAM-based interpretation.
\par
To assess LAP, we adopted two datasets from different domains and adopted domain-suitable methods for comparing interpretations. To show the general applicability of LAP, we adopted two widely used CNN architectures in our experiments, ResNet \cite{he2016deep} and Inception-V3 \cite{szegedy2016rethinking}. While both architectures have high performance, they have different core ideas. ResNet is famous for its residual connections that prepare uninterrupted gradient paths to prevent gradient fading. Inception-V3 is known for its multi-resolution analysis by applying kernels of different sizes to the feature map at each network level. The experiment setup and results for the datasets are presented in the following subsections.

\par

Based on the provided explanation, we conducted ablations on the effectiveness of the selected settings for our main contributions, e.g., various loss terms and their hyper-parameters. Both concept-discrimination loss and concordance loss enhance interpretations, but the first one results in an outstanding change. Furthermore, setting values close to the expected clue size ratio in images for MinAR and IAR hyperparameters results in a more accurate interpretation. The concept-discrimination loss approaches a simple cross-entropy using large values for these two hyperparameters. In addition, while using MaxAR has been demonstrated to be beneficial, adopting low values close to expected clue size ratios harms performance. Detailed information can be found in the supplementary material.

\subsection{RSNA pneumonia detection}
\label{subsec:rsna-celeba}

RSNA was published in a Kaggle challenge in 2018 \cite{rsna-challenge}. The dataset contains chest X-ray images of 8851 healthy people, 9555 patients having lung pneumonia, and 11821 patients with other lung abnormalities. The zones related to pneumonia have been specified by experts using bounding boxes. We used samples related to healthy people and lung-pneumonia patients in a classification task. We randomly selected 81\% of the data for training, 9\% for validation, and 10\% for test set. ResNet 18 and Inception V3 were the base architectures in these tasks. We placed three LAP modules in blocks 2, 3, and 4 of ResNet 18 and maxpool2, Mixed6a, and Mixed7a of Inception V3. The adaptive pooling was replaced with adaptive LAP in both networks.

\par
In the binary classification of "having pneumonia," the two classes have only one difference. The images related to the positive class have pneumonia in some parts, while the others do not. Because the negative images do not have any clue that positive images lack, we only need one concept head for this problem, "pneumonia." The concept head for detecting pneumonia must be partly active in positive samples and completely inactive in negative samples. There is only one concept head, so an aggregation module is unnecessary, and final attention scores are equal to the concept scores. We trained LAP-extended models with two different methods: Weak knowledge injection (WS) according to the explanations presented in \cref{subsec:ki}, and experts' knowledge injection (BB), where we used the experts' bounding boxes instead of weak supervision (Further details have been provided in supplementary material). 
\par
We evaluated the performance of the models using four metrics, accuracy, sensitivity, specificity, and balanced accuracy (BA). Sensitivity and specificity are recall factors widely used in medical domains for positive and negative classes. BA is the average of the recalls, which gives a fair metric for imbalanced datasets. The evaluation metrics on test data are presented in \cref{tab:results:rsna}. In both architectures and training schemes, the LAP-extended versions have surpassed the performance of the vanilla models.

\begin{table*}[!htb]
\centering
\caption{Results on the RSNA dataset for models trained by the proposed weakly supervised loss (WS) and expert annotations (BB). LAP extension has not resulted in a performance drop compared to the vanilla models. Accuracies of LAP predictors w.r.t. ground truth (predictivity) and model prediction (model compatibility) show their effectiveness.}
\label{tab:results:rsna}
\resizebox{\textwidth}{!}{%
\begin{tabular}{|l|cccc|cccc|cccc|}
\hline
  \multirow{2}{*}{Model} &
  \multicolumn{4}{c|}{Model performance} &
  \multicolumn{4}{c|}{LAP predictions} &
  \multicolumn{4}{c|}{Model compatibility} \\ \cline{2-13} 
  &
  Acc. &
  Sens. &
  Spec. &
  BA &
  $\text{LAP}_1$ &
  $\text{LAP}_2$ &
  $\text{LAP}_3$ &
  $\text{LAP}_4$ &
  $\text{LAP}_1$ &
  $\text{LAP}_2$ &
  $\text{LAP}_3$ &
  $\text{LAP}_4$ \\ \hline
  ResNet 18 &
  95.22 &
  95.08 &
  95.37 &
  95.23 &
  - &
  - &
  - &
  - &
  - &
  - &
  - &
  - \\
  WS-LAP ResNet 18 &
  96.58 &
  97.07 &
  96.05 &
  96.56 &
  55.48 &
  69.76 &
  88.11 &
  94.41 &
  55.86 &
  70.14 &
  88.6 &
  96.09 \\
  BB-LAP ResNet 18 &
  96.96 &
  95.92 &
  98.08 &
  \textbf{97} &
  54.06 &
  86.02 &
  95.62 &
  96.8 &
  53.96 &
  85.96 &
  96.42 &
  98.95 \\ \hline
  Inception V3 &
  96.15 &
  95.71 &
  96.61 &
  96.16 &
  - &
  - &
  - &
  - &
  - &
  - &
  - &
  - \\
  WS-LAP Inception V3 &
  96.63 &
  96.03 &
  97.29 &
  \textbf{96.66} &
  51.9 &
  51.9 &
  96.15 &
  96.25 &
  51.14 &
  51.14 &
  97.56 &
  97.67 \\
  BB-LAP Inception V3 &
  96.42 &
  96.23 &
  96.61 &
  96.42 &
  51.9 &
  93.65 &
  96.53 &
  96.36 &
  51.57 &
  94.3 &
  99.35 &
  99.73 \\ \hline
  \end{tabular}%
}
\end{table*}

\begin{figure}[!htb]
\centering
\includegraphics[width=\linewidth]{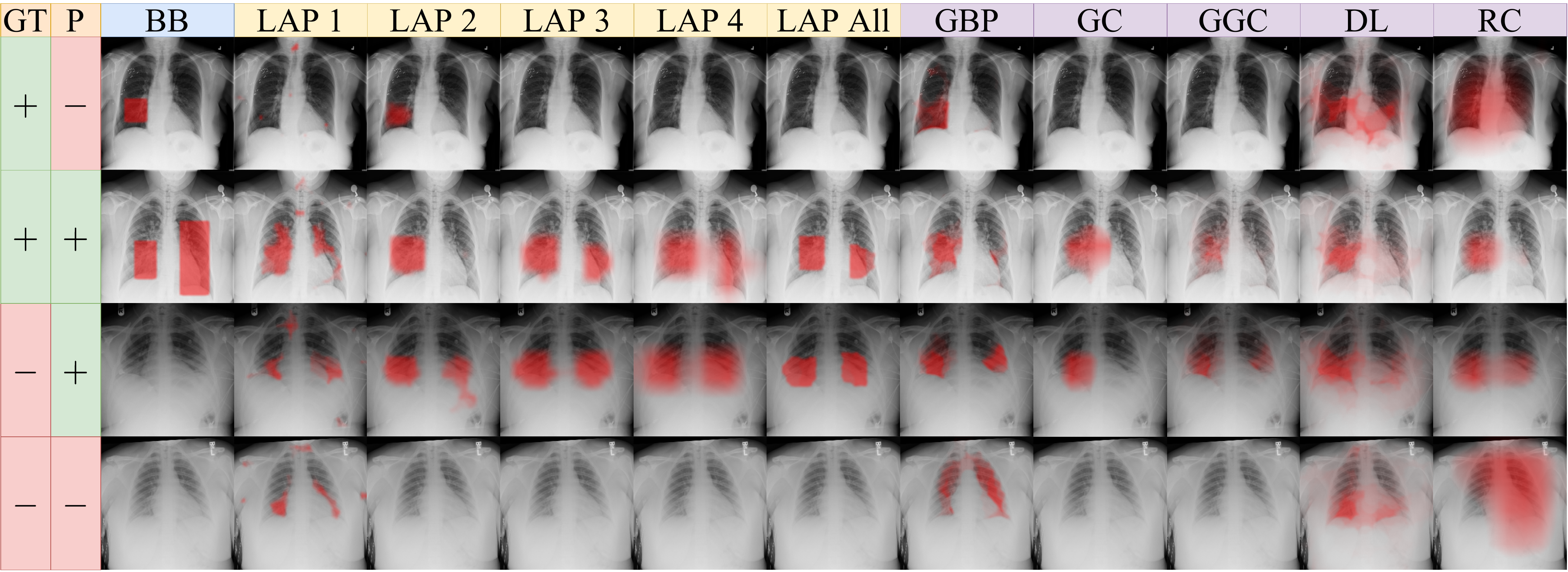}
\caption{Examples of BB-LAP Inception V3 interpretations (LAPs) compared to RSNA bounding boxes (BB) and other interpretation methods (GBP, GC, GGC, DL, and RC) on RSNA. LAP is faithful to the model's prediction (P), showing from which layer the model has differed from the ground truth (GT).}
\label{fig:inception-rsna-interpretation}
\end{figure}

Each LAP layer can also be used as a standalone predictor. In the problem of pneumonia detection, if a LAP has assigned a probability of more than 0.5 to at least one pixel, it means it has found pneumonia in the image. The prediction of the LAP for these samples is assumed to be positive and otherwise negative. We evaluated the predictivity of LAP modules as standalone deciders and the compatibility of their predictions with the model's decisions. According to \cref{tab:results:rsna}, as expected, the deeper the layer and the larger its receptive field, the higher the predictivity and the compatibility. Assessing LAPs allows us to diagnose the model from various directions. In this example, for an architecture as complicated as Inception V3, $LAP_3$ has nearly reached the performance of the whole model. Even in BB-LAP Inception V3, it has a better performance than $LAP_4$ and the final decider. We can decide that the layers after $LAP_3$ have suffered more from overfitting to the training data, resulting in a performance drop in the test data. Additionally, it can be observed that the BB-LAPs in both ResNet 18 and Inception V3 generally have higher accuracies than WS-LAPs, especially in LAP$_2$. This observation implies that using exact supervision leads to better learning in shallower layers of the network. The deep networks try to recover the performance in deeper layers. LAPs also allow us to diagnose the decision-making for one single image. The interpretation of LAPs is presented for four examples in \cref{fig:inception-rsna-interpretation}. The source of the mistakes made by the model in the first and third cases can be identified. In the first case, the $LAP_2$ layer detected the pneumonia region correctly, but the model changed its decision afterward. In the third case, the model mistakenly identified highlighted pneumonia regions from $LAP_2$. It is worth noting that the final LAP interpretation aligns with the model's decisions, whereas techniques such as RC, DL, and GBP have highlighted regions in all cases. In the true-positive decisions of the model, LAP interpretations have an acceptable overlap with infection bounding boxes, while GC, GGC, and RC have not successfully captured all boxes. Details of integrating LAPs (LAP All) and examples of the other models are provided in the supplementary.

Various evaluation metrics have been utilized in different studies to evaluate interpretation methods, each with advantages and disadvantages, making each more appropriate for specific domains. In this study, we have employed two distinct evaluation methods for our two datasets based on the nature of their domain. For the RSNA dataset, the input images are X-rays of the entire chest, and different parts of the image find different meanings based on their surroundings. In this sense, the domain differs from an object recognition dataset, where individuals can comprehend the object from its parts. Therefore, evaluation methods that involve removing parts of the image may influence the model's decision-making process due to producing out-of-distribution images. Methods like blurring may also result in artifacts that resemble pneumonia. Alternatively, the interpretation methods are expected to capture the most distinguishing parts between images that belong to different categories. All pneumonia regions are, in fact, the distinguishing parts. We have used the "object localization" \cite{lee2021relevance} to evaluate the interpretation methods. To this end, the Intersection over Union (IoU) metric was calculated between the interpretation maps and the bounding boxes provided by experts for pneumonia in true positive cases.
\par
To verify the superior performance of our self-interpretation method, we compared the interpretations with five famous white-box explainers: GC, GGC, GBP, DL, and RC. For calculating IoU, the score maps produced by interpretation methods need to be binarized. In some studies, the $k\%$ of the most highly scored pixels by the interpretation method is assumed as the final binarized interpretation mask. Methods like GC have adopted a fixed value of $15\%$ for $k$ \cite{selvaraju2017grad}. In some other studies, the binarization threshold has been calculated dynamically using the mean and standard deviation of the interpretation scores \cite{lee2021relevance}. We also adopted two different binarization methods to assess interpretations from two points of view. In the first method, rather than using a fixed percentage for all images, the number of pneumonia pixels known from the expert bounding boxes was used for each image (\textbf{Binarization by Top-Scored Selection}). We believe this results in a more fair comparison as the images in the dataset are highly diverse in the size of their pneumonia bounding boxes. The results are presented in \cref{tab:rsna-interpretation}. It is observed that LAP has achieved significantly higher performance than the other methods in all models, except in BB-LAP Inception V3, in which RC slightly outperforms LAP. Although this binarization method allows for a fair comparison of score maps by considering the size of the distinguishing area, the information is not known in test-time interpretation. Therefore, we also adopted a thresholding method for binarization to assess the interpretation maps (Binarization by Thresholding). To find a global threshold for other interpretation methods, we normalized each importance map by its maximum value. Then, we trained a binary linear classifier based on normalized scores to discriminate the pixels under bounding boxes from the others in the validation set (More details are available in the supplementary). Notably, LAPs do not need this step as they inherently provide a classification over the distinguishing pixels. The results provided in \cref{tab:rsna-interpretation} are similar to the first binarization method. The two experiments show that LAP offers a better interpretation than other methods, independent of the binarization method. Besides, LAP-provided scores do not need an external attempt for binarization.

\begin{table*}[!htb]
\centering
\caption{IoUs of infection bounding boxes of RSNA and binarized interpretation maps of different interpretation methods using two binarization approaches for four networks. LAP achieved a superior performance than other interpretation methods.}
\label{tab:rsna-interpretation}
\resizebox{\textwidth}{!}{%
\begin{tabular}{|c|cccccc|cccccc|}
\hline
Binerization Method & \multicolumn{6}{c|}{Thresholding}                              & \multicolumn{6}{c|}{Top-Scored Selection}                       \\ \hline
Model               & DL    & GGC   & GC   & GBP   & RC             & LAP            & DL    & GGC   & GC    & GBP   & RC             & LAP            \\ \hline
WS-LAP ResNet 18    & 24.37 & 15.98 & 9.34 & 19.09 & 28.49          & \textbf{36.26} & 20.44 & 15.53 & 7.14  & 15.55 & 33.02          & \textbf{44.29} \\
BB-LAP ResNet 18    & 26.18 & 16.92 & 2.2  & 18.67 & 17.75          & \textbf{46.5}  & 22.54 & 16.05 & 10.27 & 15.3  & 20.21          & \textbf{58.61} \\
WS-LAP Inception V3 & 25.56 & 22.07 & 2.04 & 25.71 & \textbf{33.35} & 31.97          & 22.94 & 20.83 & 7.82  & 20.42 & \textbf{41.52} & 40.72          \\
BB-LAP Inception V3 & 28.75 & 15.97 & 8.44 & 16.85 & 34.4           & \textbf{46.94} & 24.27 & 12.85 & 15.59 & 11.92 & 41.43          & \textbf{59.86} \\ \hline
\end{tabular}%
}
\end{table*}
\subsection{Imagenet}
In this experiment, we explored the adaptability of our LAPs in already trained models. We chose the ImageNet classification task \cite{ILSVRC15} to assess whether the LAPs can handle interpretations of objects with high variance in size. We used pre-trained ResNet 50 from the torchvision model zoo \cite{marcel2010torchvision} as the base architecture. The objects of the ImageNet dataset may be larger than the receptive field of the first three layers. Therefore, we only used a LAP in the fourth layer. To follow a general approach without requiring domain-related knowledge, we considered 1000 concept heads, each distinguishing one class from the others. Each concept head was obligated to be partly active in its respective class samples and completely inactive otherwise. We used our proposed weakly supervised loss besides the main cross-entropy loss. We first trained the LAP layer only for two epochs. Then, we fine-tuned the fourth layer (containing the LAP) and the fully connected layer of the ResNet 50 for three epochs. Further details of the training configurations and hyper-parameters are described in the supplementary. The performances of the original model, the version with its LAP trained, and the final tuned one are presented in \cref{tab:results:imagenet}. It is observable that LAP has adapted to the model while the performance is slightly improved.

\begin{table*}[!htb]
\centering
\caption{The top-1 and top-5 accuracies have been computed for the original model (ResNet 50), LAP-extended with only tuning the LAP layer (NFT), and LAP-extended with the whole layers after LAP being tuned (FT). The FT model performs slightly better than the original model and is also accommodated with self-interpretability. The accuracy of the LAP predictor w.r.t. ground truth (predictivity) and the model's prediction (model compatibility) show its effectiveness.}
\label{tab:results:imagenet}
\resizebox{\textwidth}{!}{%
\begin{tabular}{|l|cccc|cccc|cccc|}
\hline
  \multirow{2}{*}{Model} &
  \multicolumn{4}{c|}{Model performance} &
  \multicolumn{4}{c|}{LAP predictions} &
  \multicolumn{4}{c|}{Compatibility with the model} \\ \cline{2-13} 
  &
  \multicolumn{2}{c}{Top-1 Acc.} &
  \multicolumn{2}{c|}{Top-5 Acc.} &
  \multicolumn{2}{c}{Top-1 Acc.} &
  \multicolumn{2}{c|}{Top-5 Acc.} &
  \multicolumn{2}{c}{Top-1 Acc.} &
  \multicolumn{2}{c|}{Top-5 Acc.} \\ \hline
  ResNet 50 &
  \multicolumn{2}{c}{76.13} &
  \multicolumn{2}{c|}{92.86} &
  \multicolumn{2}{c}{-} &
  \multicolumn{2}{c|}{-} &
  \multicolumn{2}{c}{-} &
  \multicolumn{2}{c|}{-} \\
  LAP ResNet 50 (NFT) &
  \multicolumn{2}{c}{75.87} &
  \multicolumn{2}{c|}{92.74} &
  \multicolumn{2}{c}{-} &
  \multicolumn{2}{c|}{-} &
  \multicolumn{2}{c}{-} &
  \multicolumn{2}{c|}{-} \\
  LAP ResNet 50 (FT) &
  \multicolumn{2}{c}{\textbf{76.16}} &
  \multicolumn{2}{c|}{92.85} &
  \multicolumn{2}{c}{71.98} &
  \multicolumn{2}{c|}{86.4} &
  \multicolumn{2}{c}{71.98} &
  \multicolumn{2}{c|}{90.83} \\ \hline
\end{tabular}%
}
\end{table*}

In contrast to RSNA, which has a simple concept, ImageNet has 1000 concepts with objects of high-variance sizes. Therefore, evaluating the LAP's predictivity and model compatibility is not straightforward. We defined the concept size features, $F_C \in \mathbb{R}^{1000}$, as the sum of pixels' importance scores in each concept map. Then, we trained a 2-layer MLP network to classify the samples according to $F_C$. To evaluate the predictivity and model compatibility of the LAP, we compared the predictions of the mentioned MLP network with ground-truth and LAP-ResNet 50 (FT) predictions, respectively. The results are presented in \cref{tab:results:imagenet}. Despite the complexity of the prediction task, the high similarity between many of the classes in the domain, and the highly summarized information extracted from the concept-wise score maps, the LAP has achieved high predictivity and model compatibility.

\begin{figure*}[htb]
\centering
\includegraphics[width=\linewidth]{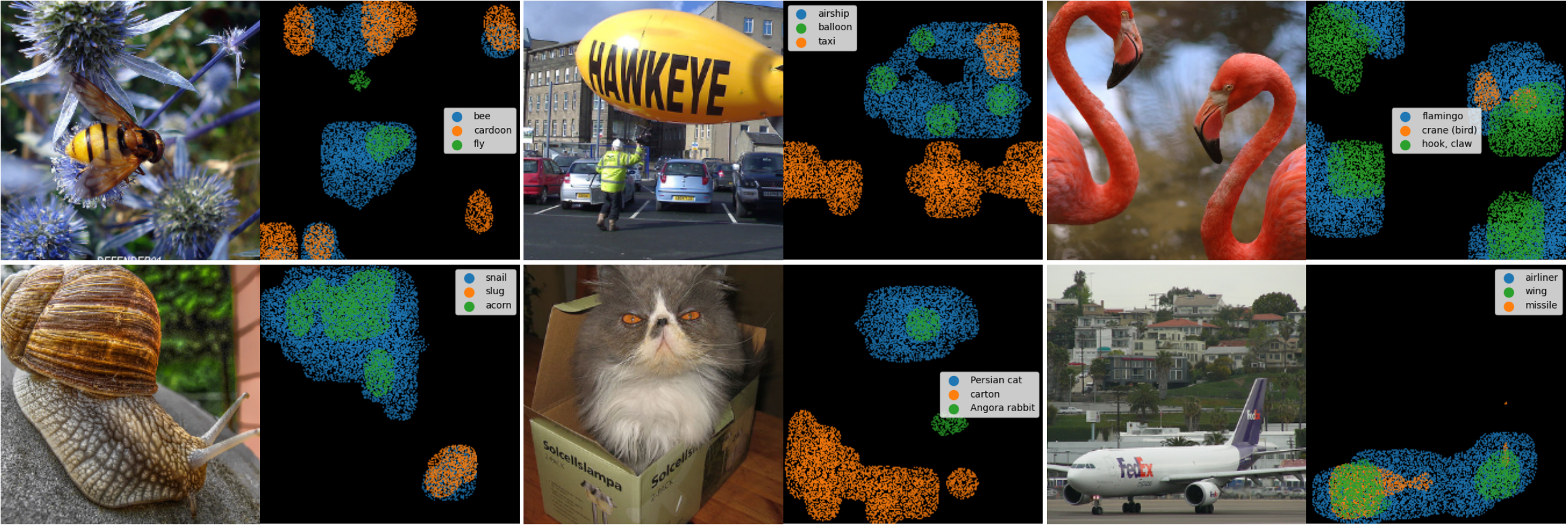}
\caption{Examples of LAP ResNet 50 interpretation on ImageNet. Different colors illustrate different concept heads. The legend shows the Top3 classes in order based on the model's predictions (More examples are provided in the supplementary).}
\label{fig:resnet-imagenet}
\end{figure*}

\begin{figure}[htb]
  \centering
  \raisebox{-\height}{\includegraphics[width=0.495\linewidth]{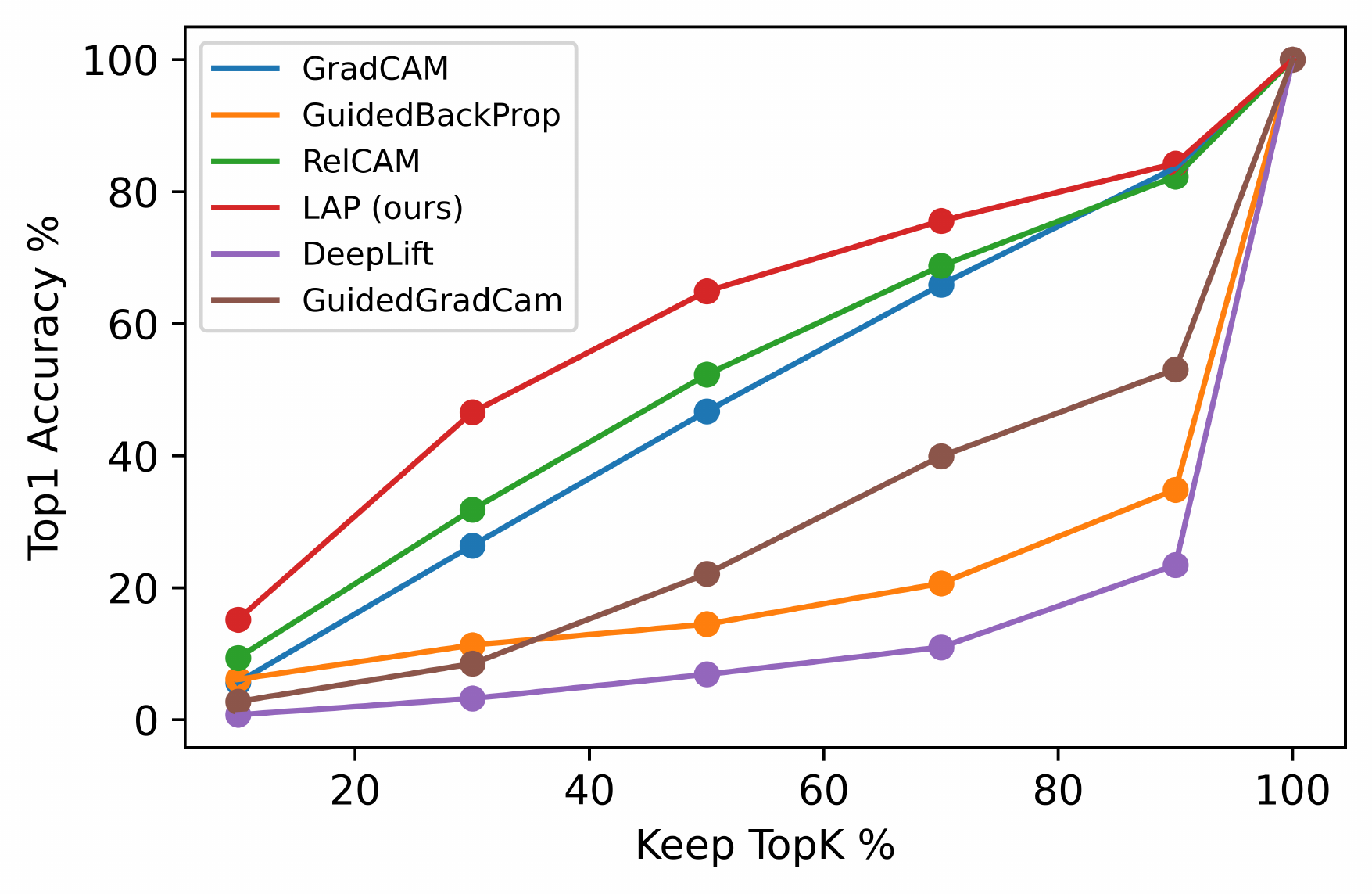}}
  \raisebox{-\height}{\includegraphics[width=0.495\linewidth]{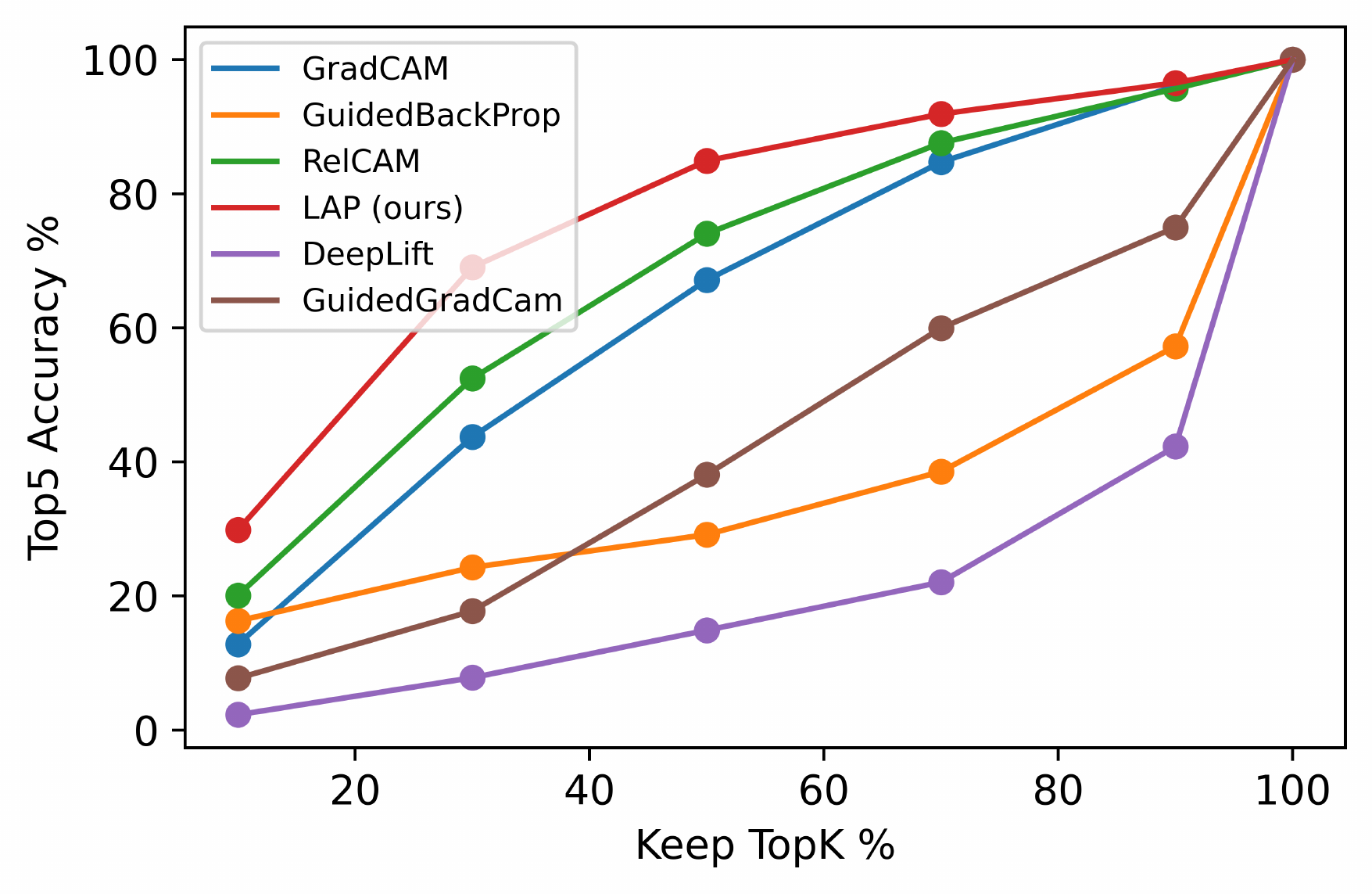}}
  \caption{Assessing the faithfulness of different interpretation methods on the ImageNet dataset. Faithfulness is evaluated by keeping a ratio of pixels with the highest scores provided by the interpretation method (K\%), then zeroing out the rest parts of the image, and evaluating the prediction based on the modified image concerning the prediction of the original image. Our method, in Top1 and Top5 accuracies, exceeds the other methods.}
  \label{fig:imagenet_f_k_pred}
\end{figure}
For comparing LAP's interpretation with other interpretation methods, we opted for a method presented in \cite{petsiuk2018rise} rather than following the same approach for the RSNA dataset. Despite the RSNA dataset in which the images of the two classes were only different in "pneumonia regions," in the ImageNet dataset, there are 1000 classes, some of which are highly similar and only differ in small details, like different classes of dogs. Therefore, the interpretation methods are not expected to highlight the whole object as distinguishing decision reasons. So, IoU cannot be a good representative of the performance. We compared the faithfulness of different interpretation methods on the tuned LAP ResNet 50 network by modifying the input images based on the interpretation scores. Faithfulness was evaluated by keeping $k\%$ of pixels with the highest scores provided by the interpretation method and zeroing out the rest of the image, then passing the modified image as input through the network for re-evaluation. The results are presented in \cref{fig:imagenet_f_k_pred} for percentages of $10\%$, $30\%$, $50\%$, $70\%$ and $90\%$. In this experiment, the interpretation scores were calculated based on the predicted class, and the results were also evaluated with respect to the original predictions. A faithful interpretation method would capture the effective regions for decision-making in the model, keeping the former decision for the modified image. Our method represents the highest faithfulness than other interpretation methods in Top1 and Top5 accuracies, introducing itself as the most confident. We also repeated this experiment based on the ground truth labels to calculate the drop-in accuracies (supplementary material). Our method reached the highest accuracy for all pixel percentages, proving it is the most faithful method to the model.
\section{Conclusion}
In this paper, we introduced the Local Attention Pool (LAP), an attention-based pooling method that can be integrated into any convolutional architecture, even if it has already been trained. We demonstrated that LAPs are capable of accommodating self-interpreting models without degrading performance. We compared LAP-based interpretation with state-of-the-art methods that did not require training a model from scratch and were applicable to any architecture. We compared interpretation maps for three models and two datasets using a technique of interpretation assessment suitable for each domain. Compared to other explainers, LAP attention maps proved more effective at faithfully explaining the model's behavior to its predictions. The LAP layers were also shown to aid in diagnosing the decision-making process of the model at different levels of the model. Additionally, they can be used as standalone predictors. Their architecture can adapt to other domains through weak or full supervision and knowledge injection. Notably, while the whole LAP architecture and proposed loss terms could also be used without plugging into the model as a pooling layer, we performed the experiments with the plugged-in version to assess task-specific performance differences. Even though LAP was found to result in some enhancement in all experiments, especially when tuning a network that had already been trained, the assumption must be validated by multiple ablation studies and repeating the experiment multiple times. Due to the purpose of the study, we did not focus on the assumption because we were attempting to demonstrate a superior ability to interpret without compromising performance. We plan to use LAPs for tasks other than classification in the future to improve explainability by addressing the issue of receptive-field dependency.

{\small
\bibliographystyle{ieee_fullname}
\bibliography{egbib}
}

\newpage
\appendix

\subsection{Fully supervised LAP loss on RSNA}
The loss function for full supervision is similar to the weakly supervised concept discrimination loss of Sec. 3.3. We used experts' annotated bounding boxes as the ground truth for active pixels for the first term. Because all the areas under a bounding box may not belong to infection zones, we applied the loss on half of the pixels with higher importance probability within the box only. All the zones out of the bounding boxes correspond to non-infection areas. We applied the second term to all of them. The third term was used similarly as the weakly supervised loss.

\subsection{Training configurations}

\subsubsection{RSNA}

\label{subsec:rsna-conf}

For weak supervision, we used cross-entropy loss on the classification head, concept-discrimination loss, and inter-LAPs concordance loss with weights of 1, 0.25 per LAP, and 0.25 per LAP pair, respectively.
The hyper-parameters of the concept-discrimination loss were as follows: $\text{MinAR} = 0.1, \text{MaxAR} = 0.5, \text{IAR} = 0.1$. 
The first two were set based on the possible range of concept sizes, i.e., infection, in the positive samples. We chose the median and maximum of the infection bounding-boxes areas as the mentioned bounds. The third was set to 0.1 to avoid fading its corresponding loss term. For full supervision, we used cross-entropy loss on the classification head alongside cross-entropy loss on LAPs, considering experts' bounding boxes as the ground truth with weights of 1 and 0.25 per LAP, respectively.

We trained the models for 300 epochs, where we observed convergence due to no change in the last 50 epochs. We selected the model related to the epoch with the best performance on validation data as the final model. We used batches of size 64 (32 healthy and 32 pneumonia samples) and the ADAM optimizer with an initial learning rate of $10^{-4}$ and a decay coefficient of $10^{-6}$ in training. The models were trained on a GEFORCE RTX 2080 Ti GPU.

\subsubsection{ImageNet}

We used our proposed weakly supervised loss (MinAR and IAR of 0.01, without MaxAR) with a factor of 0.125 beside the main cross-entropy loss. We first trained only the LAP module while other parameters were frozen for two epochs using the ADAM optimizer with an initial learning rate of $10^{-4}$ and a decay coefficient of $10^{-6}$. Then, we fine-tuned the fourth layer (containing the LAP) and the fully connected layer of the ResNet 50 for three epochs using a stochastic gradient descent optimizer with an initial learning rate of $10^{-3}$ and a decay coefficient of $10^{-6}$.

\subsection{Finding global threshold for binarizing white-box explainer's scores}
\label{sec:global-threshold}

We first normalized each importance map by its maximum value according to their papers to find a global threshold for other explainer methods. Then, we created a dataset from the normalized pixel-wise interpretation scores over the validation samples. We assigned the positive label to all the pixels under the experts' annotated boxes and the negative to the others. We used RidgeClassifier of sklearn \cite{scikit-learn} to classify the pixels. Due to the large number of pixels, we used the \textit{lsqr} solver with a tolerance of $10^{-3}$, alpha of 0.01, and set the maximum iterations to 100. We also used balanced weighting to address the issue with the highly imbalanced dataset. We used the point with the prediction label equal to zero as the threshold for binarization. We applied this method for each trained model separately and used the resulting threshold to evaluate the model's interpretations.

\subsection{Integrating LAPs' scores to one unified interpretation map}
\label{subsec:lap-aggregation}

LAPs can be considered a sequence of information through the depth of the network. Shallower LAPs cannot capture enough information due to their low receptive field. Therefore, they are likely to make more mistakes. Some pixels may have been assumed to be important but were found unimportant in the deeper layers and vice versa. However, they produce more detailed maps because of the low receptive field and high resolution. We devised an algorithm to integrate interpretations from the final LAP layers iteratively to the initial layers. In this way, we can have both accuracy and resolution. The procedure's pseudo-code is presented in \cref{alg:attention-integration}, in which $\alpha$ is the decay factor to adjust the impact of shallower LAPs. Iteratively, the algorithm modifies the current integrated map, $R_{l+1}$, with the current LAP attention map, $P_l$. Considering one pixel of $R_{l+1}$, $r_{l+1} \in \R$, and the set of its corresponding pixels in $P_l$, $p_{l} \in \R^{H_K \times W_K}$, if $r_{l+1}$ is active, i.e., greater than 0.5, at least one pixel in the corresponding zone must have been responsible. If any pixel of the $p_{l}$ is active, the credit only belongs to the active pixels. Otherwise, the current LAP has not comprehended the importance of this zone. Therefore, the credit belongs to all of them. Using this scheme, we prune the produced importance map from the last LAP, expected to be the most accurate, to the first. When choosing the topK pixels based on the ground truth bounding box size, the scores have been added without being clipped to keep the order of scores below 0.5 and have a proper selection.

\begin{algorithm}
\caption{Pseudo-code for integrating the currently integrated map pixel on the position $(i, j)$ from the $L$\textsuperscript{th} LAP to $l+1$\textsuperscript{th} LAP with $l$\textsuperscript{th} LAP attention map in the corresponding kernel. This procedure is repeated for each pixel of each LAP layer, from $L$ to $1$.}
\label{alg:attention-integration}

\begin{algorithmic}[1]

\State $\alpha \gets 0.8$ \Comment{the impact decay factor}
\State $L \gets$ The number of LAP layers
\Procedure{integratePixel}{$R, P, l, i, j, H_K, W_K$}
\State $p_l \gets \Call{getKernel}{P_l, i, j, H_K, W_K}$ \Comment{$l$\textsuperscript{th} LAP attention map for $(i, j)$'s corresponding kernel of size $H_K \times W_K$}

\If{$l = L$}
    \State \Return{$p_l$}
\EndIf

\State $r_{l + 1} \gets R_{l+1}[i, j]$ \Comment{current integrated map pixel from $L$\textsuperscript{th} LAP to $l+1$\textsuperscript{th} LAP}

\State $r_l \gets r_{l+1}$ repeated to size $H_K \times W_K$ \Comment{The result integrated map}
\If{$r_{l + 1} \geq 0.5 \And \max\{p_l\} \geq 0.5$}
    \For{$i': [0, H_K)$}
        \For{$j': [0, W_K)$}
            \State $p \gets p_l[i', j'] \times \alpha^{L-l}$ \Comment{Apply the decay factor}
            \If{$p_l[i', j'] \geq 0.5$}
                \State $r_l[i', j'] \gets \max\{r_{l+1}, p\}$
            \Else
                \State $r_l[i', j'] \gets p$
            \EndIf
        \EndFor
    \EndFor
\EndIf
\State \Return{$r_l$}
\EndProcedure
\end{algorithmic}
\end{algorithm}

\subsection{Comparing faithfulness of the interpretation methods on ImageNet with respect to the ground truth}

\begin{figure}[htb]
  \centering
  \raisebox{-\height}{\includegraphics[width=0.49\linewidth]{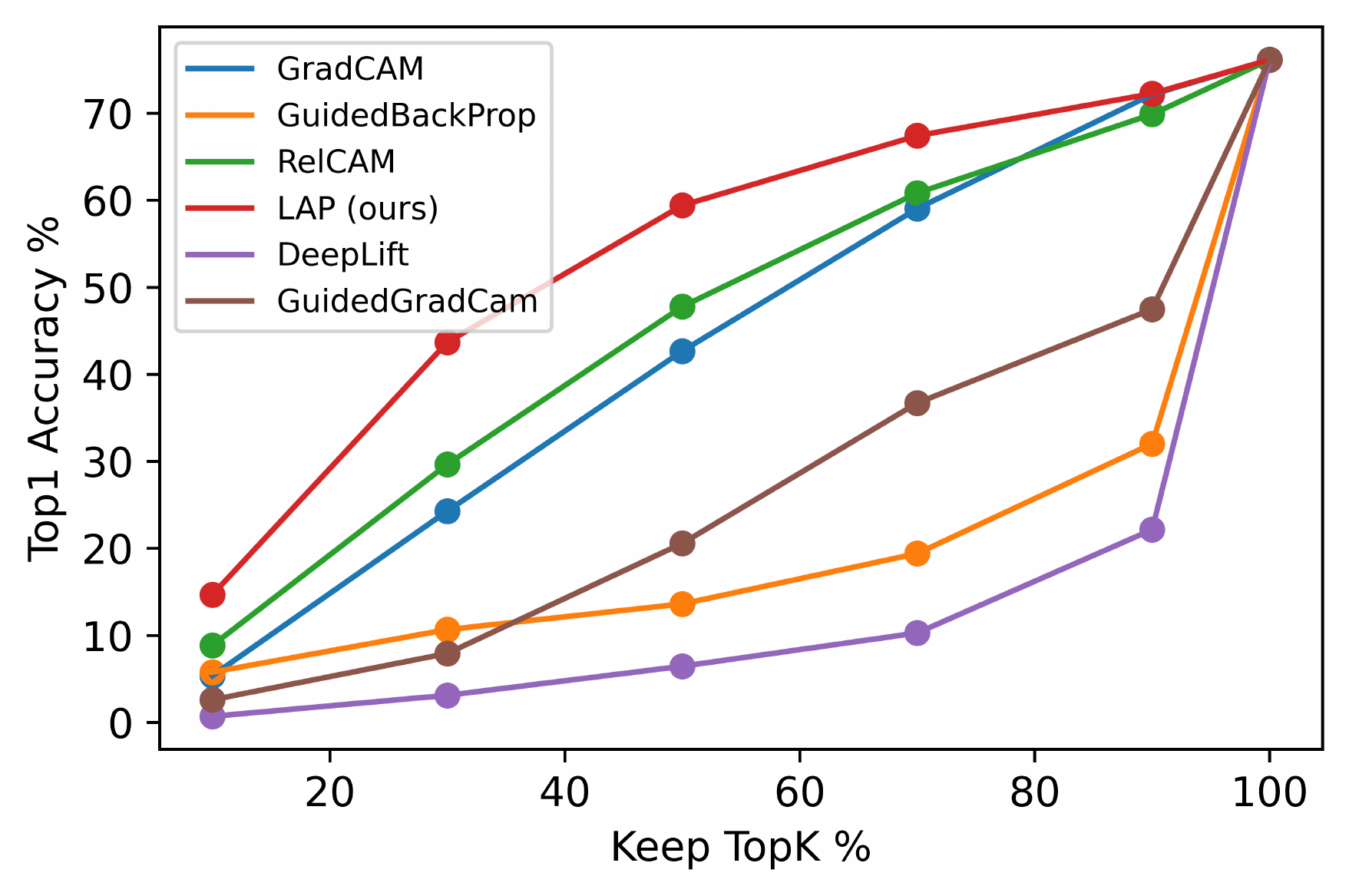}}
  \raisebox{-\height}{\includegraphics[width=0.49\linewidth]{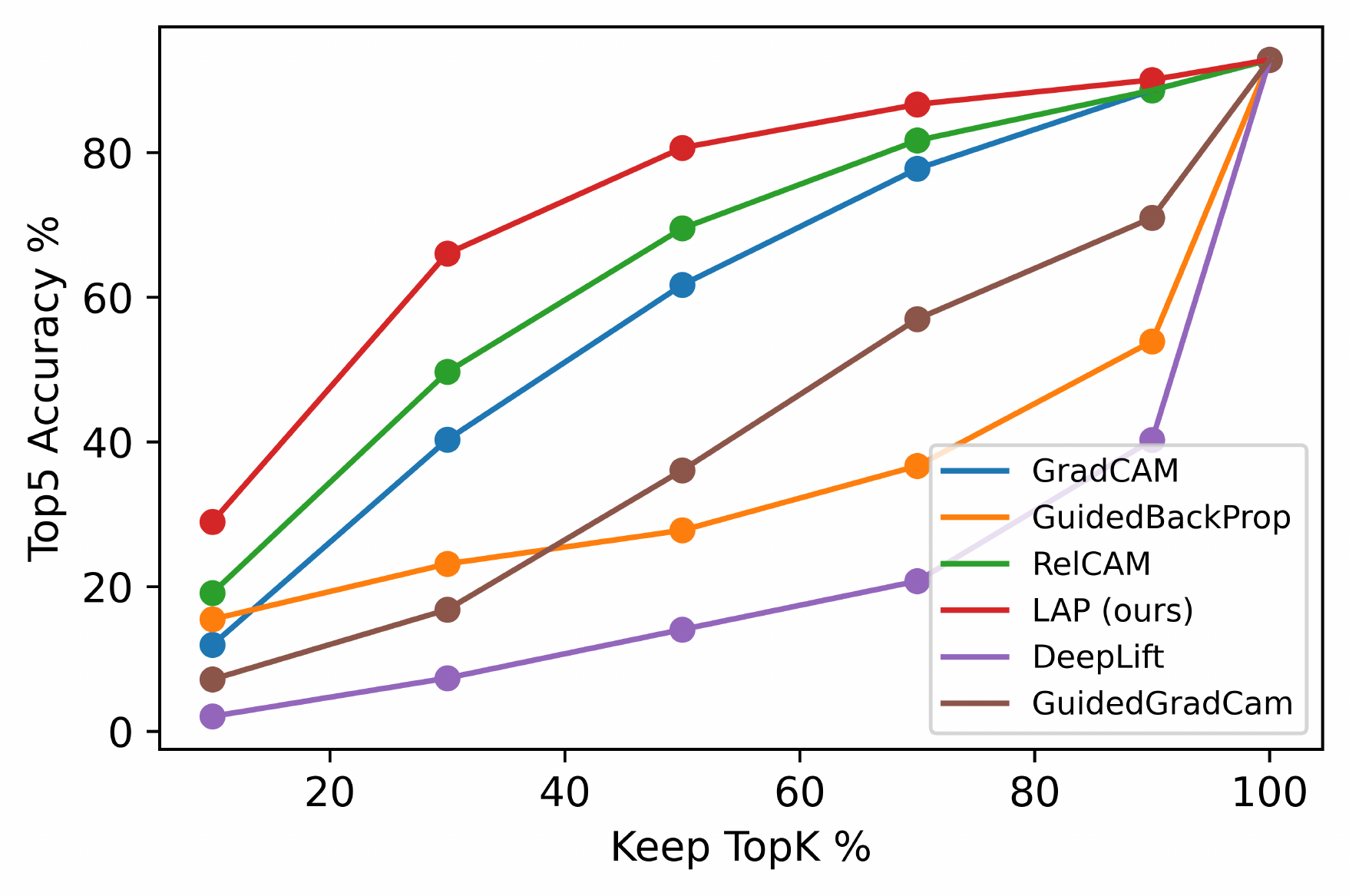}}
   \caption{Assessing the faithfulness of different interpretation methods on the ImageNet dataset. Faithfulness is evaluated by keeping a ratio of pixels with the highest scores provided by the interpretation method (K\%) for the ground truth class, then zeroing out the rest parts of the image, and evaluating the prediction based on the modified image with respect to the prediction of the original image. Our method in Top1 and Top5 accuracies exceeds the other methods.}
  \label{fig:imagenet_f_k_gt}
\end{figure}

We compared the faithfulness of different interpretation methods on the tuned LAP ResNet 50 network on the ImageNet dataset by modifying the input images based on the interpretation scores. Faithfulness was evaluated by keeping $k\%$ of pixels with the highest scores provided by the interpretation method and zeroing out the rest parts of the image, then passing the modified image as input through the network for re-evaluation. The results are presented in \cref{fig:imagenet_f_k_gt} for percentages of $10\%$, $30\%$, $50\%$, $70\%$ and $90\%$. In this experiment, the interpretation scores were calculated based on the ground truth class, and the results were also evaluated with respect to the ground truth. A faithful interpretation method would capture the effective regions for decision-making in the model, keeping the former decision for the modified image. Our method represents the highest faithfulness than other interpretation methods in Top1 and Top5 accuracies, introducing itself as the most confident.

\subsection{Ablation studies for training LAP modules}

We conducted an ablation study on the RSNA dataset using the ResNet 18 network to assess the effect of different loss terms and hyperparameters. The results are presented in \cref{tab:rsna-ablation}. In 2 settings, the network was trained without applying the concordance loss and without using any loss term to the LAP layers. The binarized interpretation maps were evaluated based on IoU with the ground truth bounding boxes. It is observed both losses have enhanced interpretations. In the case of applying no loss term to the LAPs, even though the network has been free, it has tended to pay attention to the infected regions. In 8 other settings, the network was trained with different values for MinAR, MaxAR, and IAR. The values of the hyperparameters are presented in \cref{tab:rsna-ablation-set}. It is observed that the selected setting, based on the domain's properties according to \cref{subsec:rsna-conf}, has achieved the second-highest IoU with a slight difference from the first rank. Notably, the setting achieving the first rank had the same MinAR and MaxAR selected based on the domain. The only difference was lower IAR, which had put a higher weight on the mistakes of the negative cases. Using a value much lower and much higher than the selected MinAR has affected the results in all of the proposed settings. The reason is that the chosen value, the expected size of the distinguishing part, explores the right amount of the image. A much higher value will force the network to find irrelevant pixels as distinguishing parts, which will deteriorate the gradient flow. A much lower value will prevent the network from exploring all the distinguishing parts, and it would only learn highly distinguished pixels and lose its generalization. A low value for MaxAR will also force the network to keep the distinguishing region small and cause problems in cases with large infected zones. This experiment shows that hyper-parameters can be selected according to the method described in \cref{subsec:rsna-conf}, and there is no need to try many different values. Noteworthy, all configurations but the ones related to high MinAR and all-low configuration are highly above interpretations made by other methods on the network trained by the selected values.
\begin{table}[htb]
\centering
\caption{IoUs of infection bounding boxes of RSNA and binarized interpretation maps of the LAP ResNet 18 network trained with different configurations using two binarization approaches, thresholding (TH) and topK selection by ground truth (TK). Configurations have three locations for MinAR, MaxAR, and IAR, respectively. $-$ has been used for the unchanged value, $L$ for a lower value, and $H$ for a higher value with respect to the selected configurations. \textit{No C} and \textit{No L} stand for not applying concordance loss and the whole proposed loss in LAPs, respectively. Notably, applying both loss terms has enhanced the interpretations. Additionally, the selected config based on the domain knowledge has achieved the second-highest value with a slight difference from the 1st rank, which validates the selection methodology.}
\label{tab:rsna-ablation}
\resizebox{\textwidth}{!}{%
\begin{tabular}{|c|c|cc|cccc|cccc|}
\hline
Method & --- & No C & No L & L-- & -L- & --L & LLL & H-- & -H- & --H & HHH
\\
\hline
TH & 36.26 & 30.17 & 26.32 & 30.27 & 36.03 & 37.79 & 28.64 & 25.10 & 35.12 & 33.31 & 22.03
\\
TK & 44.29 & 43.60 & 32.78 & 36.74 & 39.83 & 45.87 & 40.48 & 42.05 & 41.21 & 38.01 & 36.57
\\
\hline
\end{tabular}%
}
\end{table}

\begin{table}[htb]
\centering
\caption{The values used for MinAR, MaxAR, and IAR hyperparameters in the ablation study of RSNA.}
\label{tab:rsna-ablation-set}
{%
\begin{tabular}{|c|ccc|}
\hline
Type & - & L & H
\\
\hline
MinAR & 0.1 & 0.01 & 0.5
\\
MaxAR & 0.5 & 0.1 & 0.9
\\
IAR & 0.1 & 0.01 & 0.5
\\
\hline
\end{tabular}%
}
\end{table}

\subsection{More examples of LAP interpretations}
\label{sec:more-examples}

\begin{figure}[htb]
    \centering
    \includegraphics[width=\linewidth]{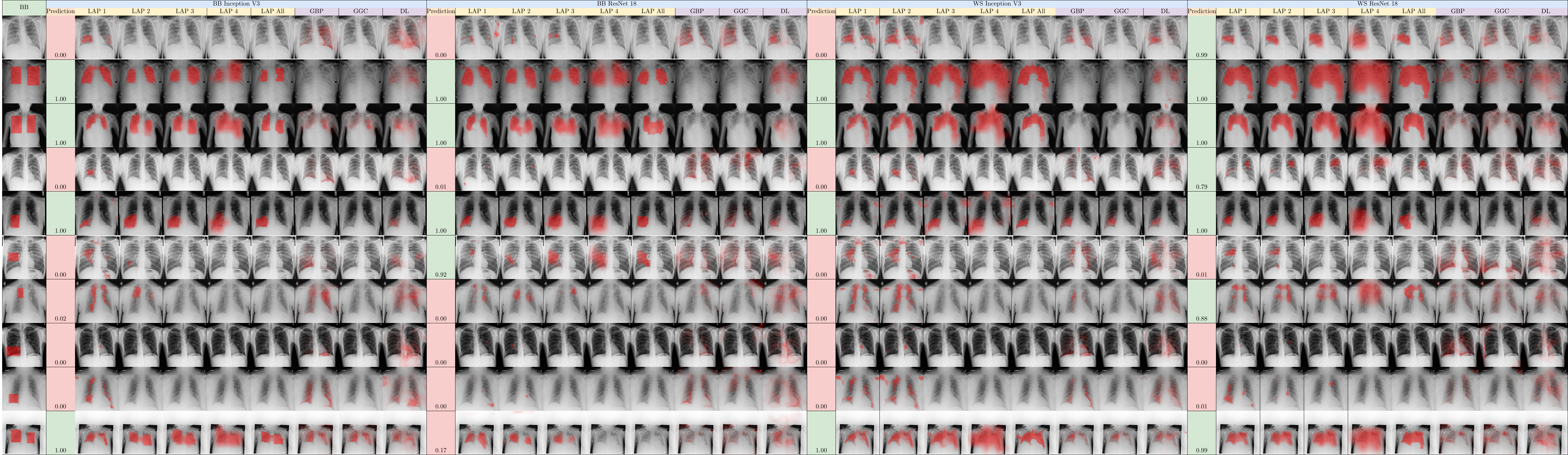}
    \caption{Examples of all models' interpretations (LAPs) compared to RSNA bounding boxes (BB) and other interpretation methods (GBP, GC, GGC, DL, and RC) on the RSNA dataset. The probability assigned by each model is presented, and the background color represents the final decision of the model. Despite other methods, LAP interpretation is faithful to the model's prediction, and it helps diagnose the layer from which the model has made a mistake in decision-making.}
    \label{fig:rsna}
\end{figure}

\begin{figure*}[htb]
    \centering
    \begin{tabular}{rl}
        \includegraphics[width=0.47\textwidth]{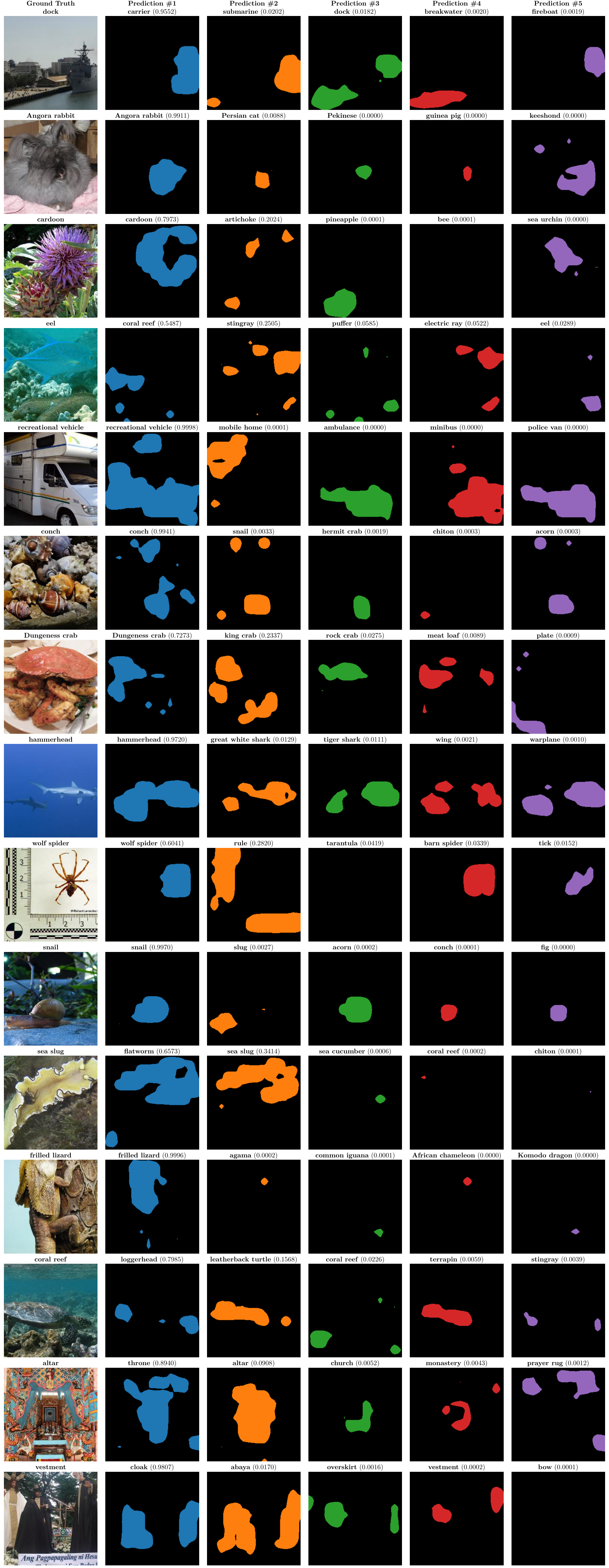}
        &
        \includegraphics[width=0.47\textwidth]{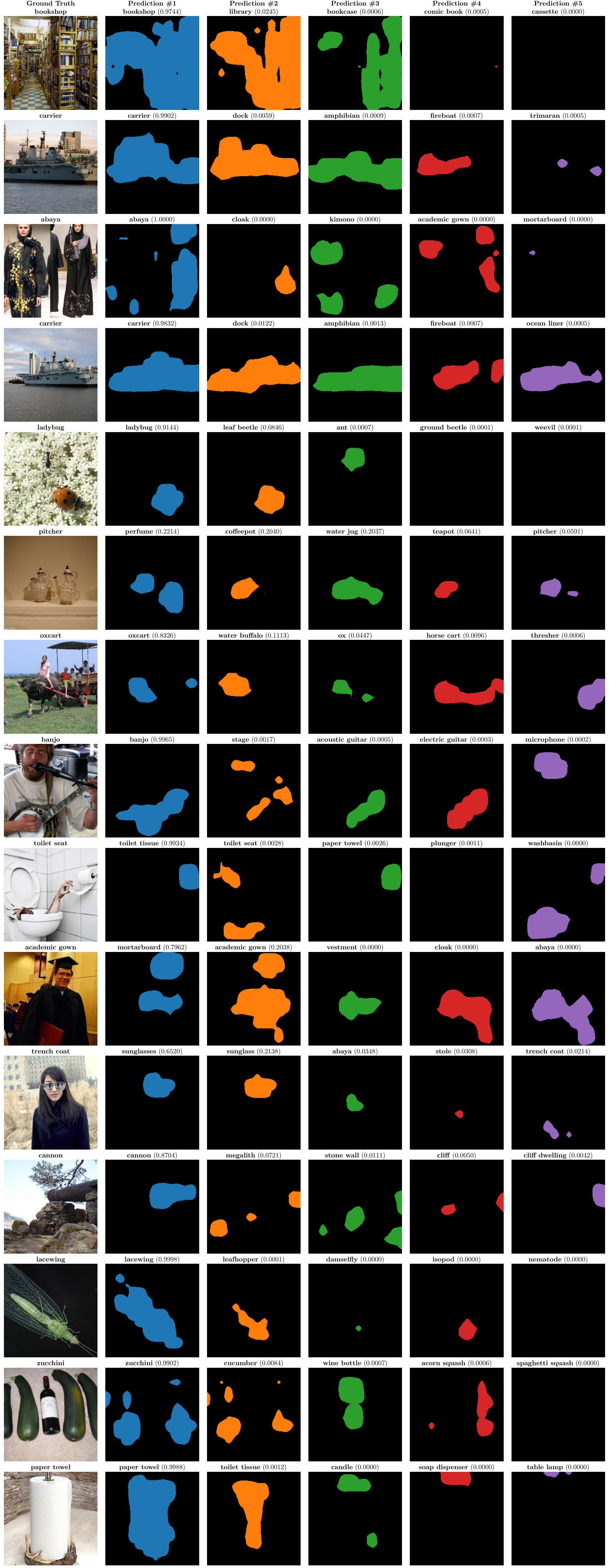}
    \end{tabular}
    \caption{Further Concept-wise interpretation Examples of LAP ResNet 50 on ImageNet dataset. For each example, concept heads related to the top-5 predictions of the model are illustrated separately. The probability assigned by the model for each concept head is presented too. The interpretation maps clarify why the model has chosen each class as one of its top 5.}
    \label{fig:imagenet}
\end{figure*}

Due to the limited number of pages in the main paper, we have provided more images interpreted with LAP for RSNA and Imagenet in \cref{fig:rsna,,fig:imagenet}, respectively.

\subsection{Licenses of the assets}

\begin{itemize}

    \item \textbf{RSNA}: The dataset is freely available for download from \href{https://www.kaggle.com/competitions/rsna-pneumonia-detection-challenge/}{Kaggle}. Based on \href{https://www.kaggle.com/competitions/rsna-pneumonia-detection-challenge/rules}{the competition website} "the competition data is allowed to be used for the competition, participation on Kaggle website forums, academic research and education, and other commercial or non-commercial purposes as long as the attribution for the dataset and the individual items (sound files) are provided when required."
    
    \item \textbf{ImageNet}: The dataset of ImageNet Large Scale Visual Recognition Challenge 2012-2017 is freely available for download from \href{https://www.image-net.org/challenges/LSVRC/index.php}{the ImageNet website}. Based on \href{https://www.image-net.org/download.php}{the website}, "researchers shall use the Database only for non-commercial research and educational purposes."
    
\end{itemize}

\end{document}